\documentclass[a4paper]{article}
\usepackage{INTERSPEECH_v2}
\usepackage{amsmath}
\usepackage{amsfonts}
\usepackage{amssymb}
\usepackage{epstopdf}
\usepackage{enumerate}

\title{Structured-based Curriculum Learning for\\End-to-end English-Japanese Speech Translation}
\name{Takatomo Kano, Sakriani Sakti, Satoshi Nakamura}
\address{Graduate School of Information Science, Nara Institute of Science and Technology, Japan}
\email{\{kano.takatomo.km0,ssakti,s-nakamura\}@is.naist.jp}

\begin{document}

\maketitle
\begin{abstract}
 Sequence-to-sequence attentional-based neural network architectures have been shown to provide a powerful model for machine translation and speech recognition. Recently, several works have attempted to extend the models for end-to-end speech translation task. However, the usefulness of these models were only investigated on language pairs with similar syntax and word order (e.g., English-French or English-Spanish). In this work, we focus on end-to-end speech translation tasks on syntactically distant language pairs (e.g., English-Japanese) that require distant word reordering.
 To guide the encoder-decoder attentional model to learn this difficult problem, we propose a structured-based curriculum learning strategy.
 Unlike conventional curriculum learning that gradually emphasizes difficult data examples, we formalize learning strategies from easier network structures to more difficult network structures. Here, we start the training with end-to-end encoder-decoder for speech recognition or text-based machine translation task then gradually move to end-to-end speech translation task. The experiment results show that the proposed approach could provide significant improvements in comparison with the one without curriculum learning.
\end{abstract}

\vspace{-0.1cm}
\section{Introduction}
\vspace{-0.1cm}

Translating a spoken language, in other words recognizing speech and automatically having one’s words translated into another language, is extremely complex. One traditional approach in speech-to-text translation systems must construct automatic speech recognition (ASR) and machine translation (MT) system, both of which are independently trained and tuned. Given a speech input, the ASR system processes and transforms the speech into the text in the source language, and then MT transforms the text in the source language to corresponding text in the target language \cite{Satoshi-IEEE06}. The basic unit for information sharing between these components is only words at the text level. Even though significant progress has been made and various commercial speech translation systems have been introduced, this approach continues to suffer from several major limitations.

One of the drawbacks is that speech acoustics might involve both linguistic and paralinguistic information (i.e., prosody, intonation, accent, rhythm, emphasis, or emotion), but such paralinguistic information is not a factor in written communication, and much cannot even be expressed in words. Consequently, the words output by ASR have lost all of their paralinguistic information, and only the linguistic parts are translated by the MT system. Some studies have proposed including additional component to just handle paralinguistic translation, but this step introduces more complexity and delay \cite{Kano-InterSpeech13,Truong-InterSpeech16,1660081}. Another noted problem is that over half of the world's languages actually have no written form and are only spoken. Another solution is to translate directly from phoneme-based transcription. However, the performance of a phoneme-based ASR is usually low, and errors in the ASR stage can propagate throughout the translation process \cite{Deng-ICASSP11}. Therefore, it would be useful to find ways beyond the current conventional approach to directly translate from the speech of the source language to the text of the target language.

Recently, deep learning has shown much promise in many tasks. A sequence-to-sequence attention-based neural network is one architecture that provides a powerful model for machine translation and speech recognition \cite{Chorowski-CoRR15,Bahdanau-CoRR14}. Recently, several works have extended models for end-to-end speech translation (ST) tasks. Duong et al. \cite{Duong-NAACL16}. directly trained attentional models on parallel speech data. But their work is only applicable for Spanish-English language pairs with similar syntax and word order (SVO-SVO). Furthermore, it focused on alignment performance. The only attempt to build a full-fledged end-to-end attentional-based speech-to-text translation system is B$\acute{e}$rard et al. \cite{Alexandre-NIPS16}. But, that work was only done on a small French-English synthetic corpus, because these language share similar word order (SVO-SVO). For such languages, only local movements are sufficient for translation.

This paper proposes a first attempt to build an end-to-end attention-based ST system on syntactically distant language pairs that suffers from long-distance reordering phenomena. We train the attentional model on English-Japanese language pairs with SVO versus SOV word order. To guide the encoder-decoder attentional model to learn this difficult problem, we proposed a structured-based curriculum learning strategy. Unlike the conventional curriculum learning that gradually emphasize difficult data examples, we formalize CL strategies that start the training with an end-to-end encoder-decoder for speech recognition or text-based machine translation tasks and gradually train the network for end-to-end speech translation tasks by adapting the decoder or encoder parts. Here we start the training with an end-to-end encoder-decoder for speech recognition or a text-based machine translation task and gradually move to an end-to-end speech translation task.

\vspace{-0.1cm}
\section{Related Works}
\vspace{-0.1cm}

\textit{Curriculum learning}, which is one learning paradigm, is inspired by the learning processes of humans and animals that learn from easier aspects and gradually increase to more difficult ones. Although the application of such training strategies to machine learning has been discussed between machine learning and cognitive science researchers going back to Elman (1993) \cite{Elman-Cognition93}, CL's first formulation in the context of machine learning was introduced by Bengio et al. (2009) \cite{Bengio-09}.

Using CL might help avoid bad local minima and speed up the training convergence and improve generalization. These advantages have been empirically demonstrated in various tasks, including shape recognition \cite{Bengio-09}, object classification \cite{Gong-IEEE16}, and language modeling tasks \cite{Rush-CoRR15}. However, most studies focused on how to organize the sequence of the learning data examples in the context of single task learning. Bengio at al. \cite{Bengio-09} proposed curriculum learning for multiple tasks. But again, all of the tasks still belonged to the same type of problem, which is object classification, where those tasks shared the same input and output spaces.

In contrast with most previous CL studies, (1) we utilize CL strategy not for simple recognition/classification problems, but for sequence-to-sequence based neural network learning problems in speech translation tasks; (2) the attentional-based neural network is not trained directly for the speech translation task using similar but more and more difficult speech translation data. Instead we formalize CL strategies that start the training with an end-to-end encoder-decoder for speech recognition or text-based machine translation tasks and gradually train the network
for end-to-end speech translation tasks by adapting the decoder or encoder parts respectively; (3) those different tasks of speech recognition, text-based machine translation, and speech translation used in structured-based CL do not share the same input and output spaces, as in the CL of multiple tasks.

\begin{figure*}
  \centering
  \fbox{\includegraphics[width=13cm]{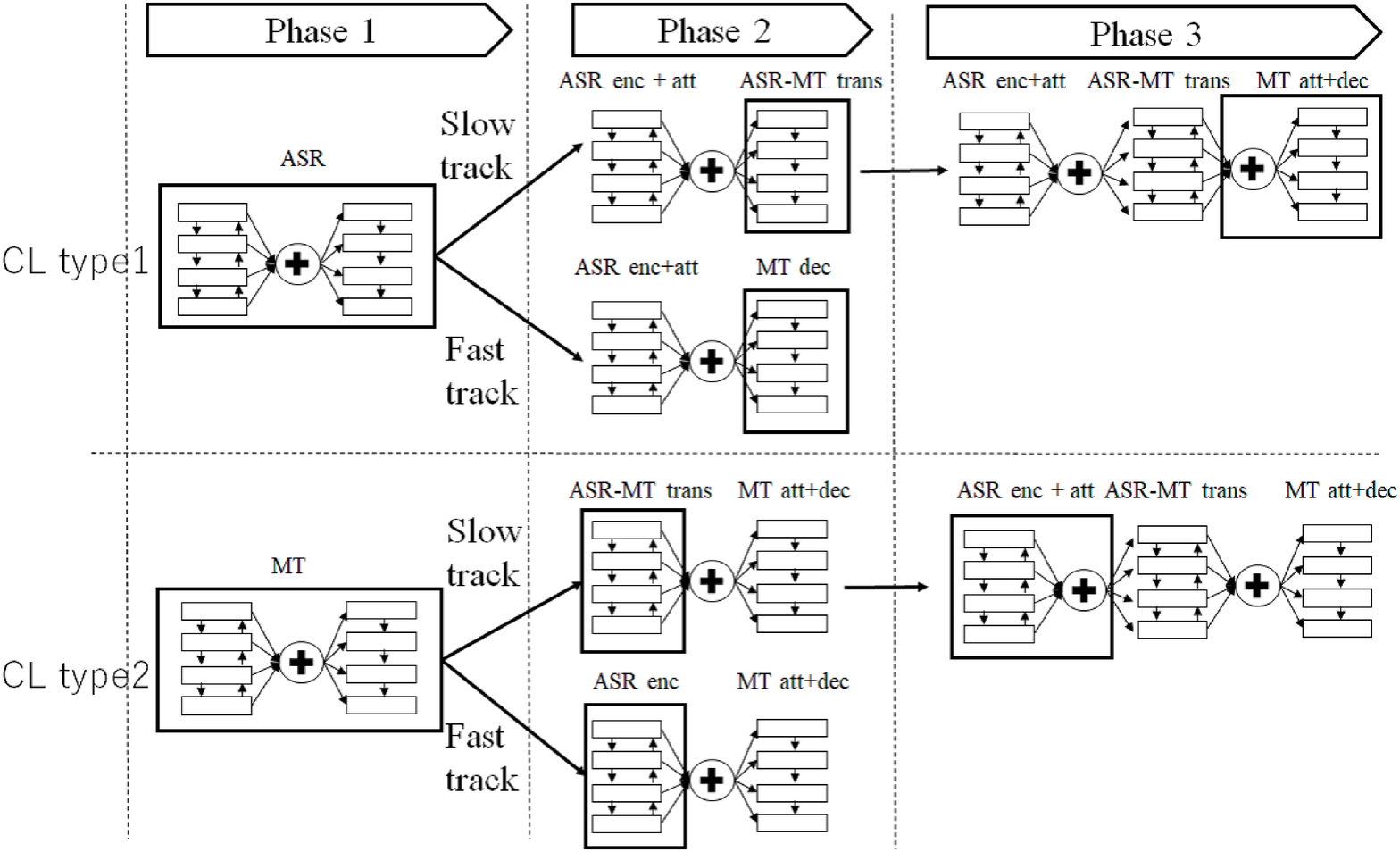}}
  \caption{Attention-based speech translation training phases with CL-based concept.}
  \label{fig1}
\end{figure*}

\vspace{-0.1cm}
\section{Basic Attention-based Speech Translation}
\label{sec:basic}
\vspace{-0.1cm}

We built our end-to-end speech translation system upon the standard attention-based encoder-decoder neural networks architecture \cite{Bahdanau-CoRR14,Sutskever-CoRR14} that consists of encoder, decoder, and attention modules.
Given input sequence $\mathbf{x} = [x_1, x_2, ..., x_N]$ with length $N$, the encoder produces a sequence of vector
representation $h^{enc} = \left(h^{enc}_1,h^{enc}_2, ..., h^{enc}_N\right)$. Here we used a bidirectional recurrent
neural network with long short-term memory (bi-LSTM) units \cite{Hochreiter:1997:LSM:1246443.1246450}, which consist of forward and backward LSTMs. The forward
LSTM reads the input sequence from $x_1$ to $x_N$ and estimates forward $\overrightarrow{h^{enc}}$, while the backward LSTM
reads the input sequence in reverse order from $x_N$ to $x_1$ and estimates backward $\overleftarrow{h^{enc}}$. Thus,
for each input $x_n$, we obtain $h^{enc}_n$ by concatenating forward $\overrightarrow{h^{enc}}$ and backward
$\overleftarrow{h^{enc}}$.

The decoder, on the other hand, predicts target sequence $\mathbf{y} = [y_1,y_2, ..., y_T]$ with length $T$
by estimating conditional probability $p(\mathbf{y}|\mathbf{x})$. Here, we use uni-directional LSTM (forward only).
Conditional probability $p(\mathbf{y}|\mathbf{x})$ is estimated based on the whole sequence of the previous output:
\vspace{-0.1cm}
\begin{equation}
p(y_t|y_1,y_2,...,y_{t-1},x)=softmax(W_y \tilde{h}^{dec}_t).
\end{equation}
Decoder hidden activation vector $\tilde{h}^{dec}_t$ is computed by applying linear
layer $W_c$ over context information $c_t$ and current hidden state $h^{dec}_t$:
\begin{equation}
\tilde{h}^{dec}_t = \text{tanh}(W_c[c_t;h^{dec}_t]).
\end{equation}

Here $c_t$ is the context information of the input sequence when generating current output at time $t$, estimated by the attention module
over encoder hidden states $h^{enc}_n$
\begin{equation}
c_t = \sum_{n=1}^{N} a_t(n) * h^{enc}_n,
\end{equation}
where variable-length alignment vector $a_t$, whose size
equals the length of input sequence $x$, is computed by
\begin{eqnarray}
  a_t(n) &=& \text{align}({h^{enc}_n}, h^{dec}_t) \\ \nonumber
  &=& \text{softmax}(\text{dot}({h^{enc}_n}, h^{dec}_t).
\end{eqnarray}
This step is done to assist the decoder to find relevant information on the encoder side based on the current decoder hidden states.
There are several variations to calculate $\text{align}({h^{enc}_n}, h^{dec}_t)$. Here we simply use the dot product
between the encoder and decoder hidden states \cite{LuongPM15}.

In this study, we apply this basic architecture for various tasks:
\begin{itemize}
\item ASR system\\
Input sequence $\mathbf{x} = [x_1, ..., x_N]$ is the input speech sequence of the source language, and target
sequence $\mathbf{y} = [y_1, ..., y_T]$ is the predicted corresponding transcription in the source language.
\item MT system\\
Input sequence $\mathbf{x} = [x_1, ..., x_N]$ is the word sequence of the source language, and target
sequence $\mathbf{y} = [y_1, ..., y_T]$ is the predicted corresponding word sequence in the target language.
\item ST system\\
Input sequence $\mathbf{x} = [x_1, ..., x_N]$ is the input speech sequence of the source language, and target
sequence $\mathbf{y} = [y_1, ..., y_T]$ is the predicted corresponding word sequence in the target language.
\end{itemize}

\vspace{-0.1cm}
\section{Attention-based Speech Translation with Curriculum Learning}
\label{sec:proposed}
\vspace{-0.1cm}

The training process of the attention-based encoder-decoder model is basically more difficult than the standard neural network model \cite{chan2016listen} because an attention-based model needs to jointly optimize three different (encoder, decoder, and attention) modules simultaneously. Utilizing the attention-based encoder-decoder architecture for constructing a direct ST task is obviously difficult because the model needs to solve two complex problems: (1) learning how to process a long speech sequence and map it to the corresponding words, similar to the issues focused on in the field of ASR \cite{Chorowski-CoRR15}; (2) learning how to make good alignment rules between source and target languages, similar to the issues discussed in the field of MT \cite{Bahdanau-CoRR14, Koehn-NAACL03}. Furthermore, we utilize attention-based encoder-decoder architecture to construct a ST system on syntactically distance language pairs that suffer from long-distance reordering phenomena and train the attentional model on English-Japanese language pairs with SVO versus SOV word order. Therefore, to assist the encoder-decoder model to learn this difficult problem, we proposed a structured-based curriculum learning strategy.

In our CL strategy, the attentional-based neural network is not trained directly for speech translation tasks using
similar but more and more difficult speech translation data, instead we formalize structured-based CL strategies that start the training
with an end-to-end encoder-decoder for ASR or text-based MT tasks and gradually train the network for end-to-end ST tasks. In other words, we train the attentional encoder-decoder architecture by starting from a simpler task, switch a certain part of the structure (encoder or decoder) in each training phase, and set it to a more difficult target task. In this way, the difficulty of the problems increases gradually in each training phase, as in CL strategies.

Figure \ref{fig1} illustrates the attention-based speech translation training phases, and the details are described below.
\begin{enumerate}
\item \textbf{CL type 1: Start from an attention-based ASR system}
Here the curriculum learning for each phases is designed as follows:
\begin{enumerate}[(a)]
\item \textbf{Fast track}
\begin{description}
\item[Phase 1] We train an attentional-based encoder-decoder neural network for a standard ASR task, which predicts the corresponding transcription of the input speech sequence in the source language.
\item[Phase 2] Next we replace the ASR decoder with a new decoder and retrain it to match the MT decoder's output. The model now predicts the corresponding word sequence in the target language given the input speech sequence of the source language.
\end{description}
\item \textbf{Slow track}
\begin{description}
\item[Phase 1] As before, we train the attentional-based encoder-decoder neural network for a standard ASR task, which predicts the corresponding transcription of the input speech sequence in the source language.
\item[Phase 2] Then we replace the ASR decoder with a new decoder and retrain it to match the MT encoder's output this work as ASR-MT transcoder. The model's objective now is to predict the word representation (like the MT encoder's output) that is good for the corresponding word sequence in the source language given the input speech sequence of the source language. Here, as a loss function, we calculate the mean squared error between the
    output of the new decoder with the ouput of the MT encoder.
\item[Phase 3] Finally, we combine the MT attention and decoder modules to perform the speech translation task from the source speech sequence to the target word sequence and train the whole architecture using a softmax cross-entropy function.
\end{description}
\end{enumerate}
\item \textbf{CL type 2: Start from attention-based MT system}
Similar to CL type 1, we construct an attentional-based ST system for both fast and slow tracks, but instead of starting with an ASR system, we start with the MT system. In this case, the model gradually adapts the encoder part from the MT encoder to more closely resemble the ASR encoder.
\end{enumerate}

\vspace{-0.1cm}
\section{Experimental Set-Up and Results}
\vspace{-0.1cm}
\subsection{Experimental Set-Up}
\vspace{-0.1cm}
We conducted our experiments using a basic travel expression corpus (BTEC)  \cite{BTEC,1677987}. The BTEC English-Japanese parallel corpus consists of 4.5-k training sentences and 500 sentences in the test set. Since corresponding speech utterances for this text corpus are unavailable, we used the Google text-to-speech synthesis\footnote{Google TTS: https://pypi.python.org/pypi/gTTS} to generate a speech corpus of the source language.

The speech utterances were segmented into multiple frames with a 25-ms window size and a 10-ms step size. Then we extracted 23-dimension filter bank features using Kaldi's feature extractor \cite{Povey_ASRU2011_2011} and normalized them to have zero mean and unit variance. As for the text corpus, using one-hot vectors results in large sparse vectors due to a large vocabulary. In this study, we incorporated word embedding that learns the dense representation of words in a low-dimensional vector space.

We further used this data to build an attention-based ASR and MT system, a direct ST system, and a CL-based ST-system. Table \ref{tb:table}
summarizes the network parameters. For all the systems, we used the same learning rate and adopted Adam\cite{DBLP:journals/corr/KingmaB14} to all of the models.

\begin{table}[h]
  \center
  \footnotesize
  \begin{tabular}{|c|c|} \hline
    \multicolumn{2}{|c|}{ASR system}\\ \hline
    Input units & 23 \\ \hline
    Hidden units & 512 \\ \hline
    Output units & 27293 \\ \hline
    LSTM layer depth & 2 \\ \hline \hline \hline
    \multicolumn{2}{|c|}{MT system}\\ \hline
    Source vocabulary  & 27293 \\ \hline
    Target vocabulary & 33155 \\ \hline
    Embed size & 128 \\ \hline
    Input units & 128 \\ \hline
    Hidden units & 512 \\ \hline
    Output units & 33155 \\ \hline
    LSTM layer depth & 2 \\ \hline\hline\hline
    \multicolumn{2}{|c|}{Optimization}\\ \hline
    Initial learning rate & 0.001000 \\ \hline
    Learning descend rate & 1.800000 \\ \hline \hline
    Optimizing method & Adam \cite{DBLP:journals/corr/KingmaB14}\\ \hline
   \end{tabular}
   \caption[t]{Model settings for each system}
  \label{tb:table}
\end{table}

\subsection{Results and Discussion}
We applied the attentional encoder-decoder architecture described in Section \ref{sec:basic} to train the ASR, MT, and direct ST systems. We also constructed an ASR+MT cascade system. For our proposed models, we also applied the CL-based attentional encoder-decoder architecture described in Section \ref{sec:proposed} to train CL type 1 and CL type 2 (fast and slow tracks). Unfortunately, CL type 2 failed to converge. This might be due to the large divergence between the MT encoder in the text input space to the ASR encoder in the speech input space. The successfully trained systems are listed below.
\begin{description}
\item[Baseline ASR:] speech-to-text model of source language.
\item[Baseline MT:] text-to-text translation model from source language to target language.
\item[Baseline ASR+MT:] speech-to-text translation model by cascading speech-to-text in source language with a text-to-text translation model.
\item[Direct ST Enc-Dec:] direct end-to-end speech translation model using a single attention-based neural network.
\item[Proposed ST Enc-Dec (CL type 1 - Fast Track):] end-to-end speech translation model trained using CL type 1 (fast track).
\item[Proposed ST Enc-Dec (CL type 1 - Slow Track):] end-to-end speech translation model trained using CL type 1 (slow track).
\end{description}

The performance of our ASR system achieved a 9.4$\%$ word error rate (WER). The remaining systems were evaluated based on translation quality using a
standard automatic evaluation metric BLEU+1 \cite{lin-och:2004:COLING}.

\begin{figure}
  \centering
  \fbox{\includegraphics[width=7.8cm]{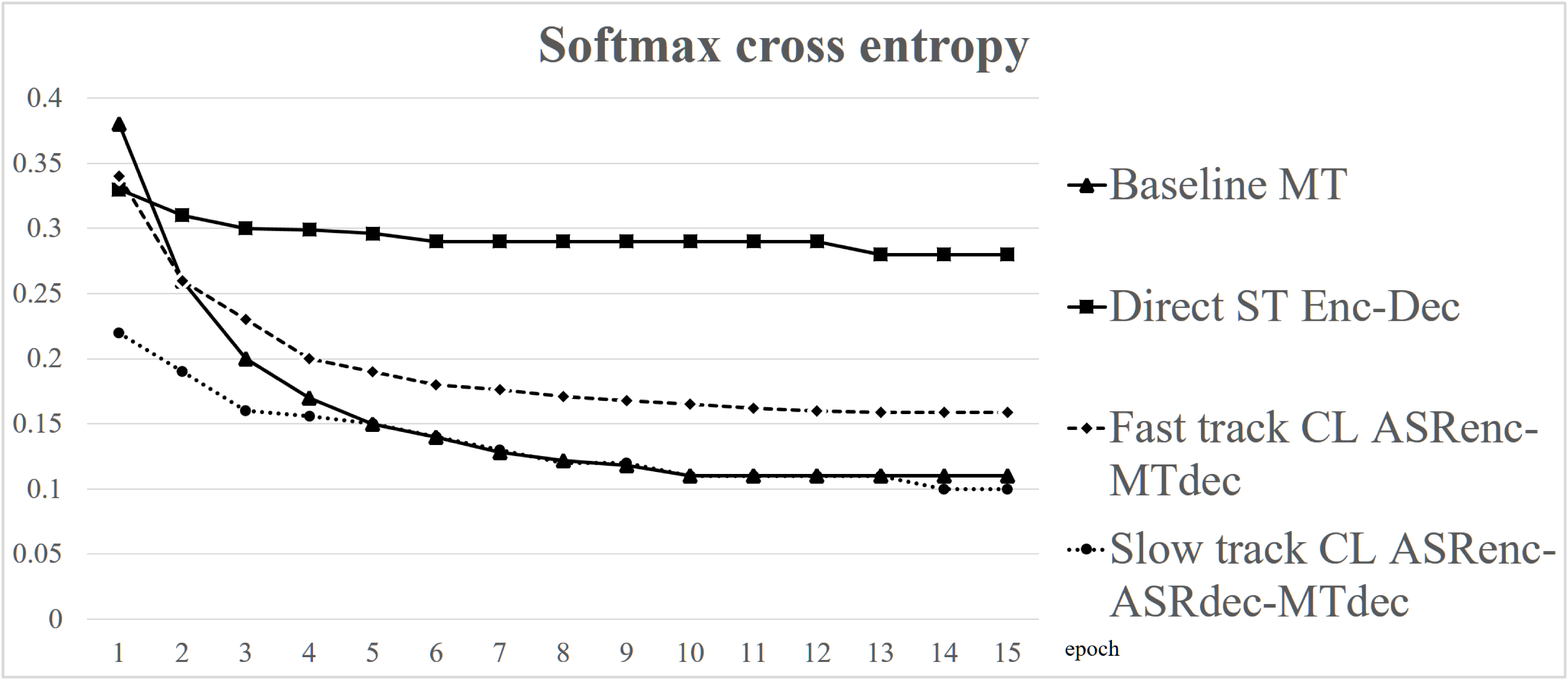}}
  \vspace{-0.3cm}
  \caption{Softmax cross-entropy of each epoch}
  \label{loss}
  \centering
  \fbox{\includegraphics[width=7.8cm]{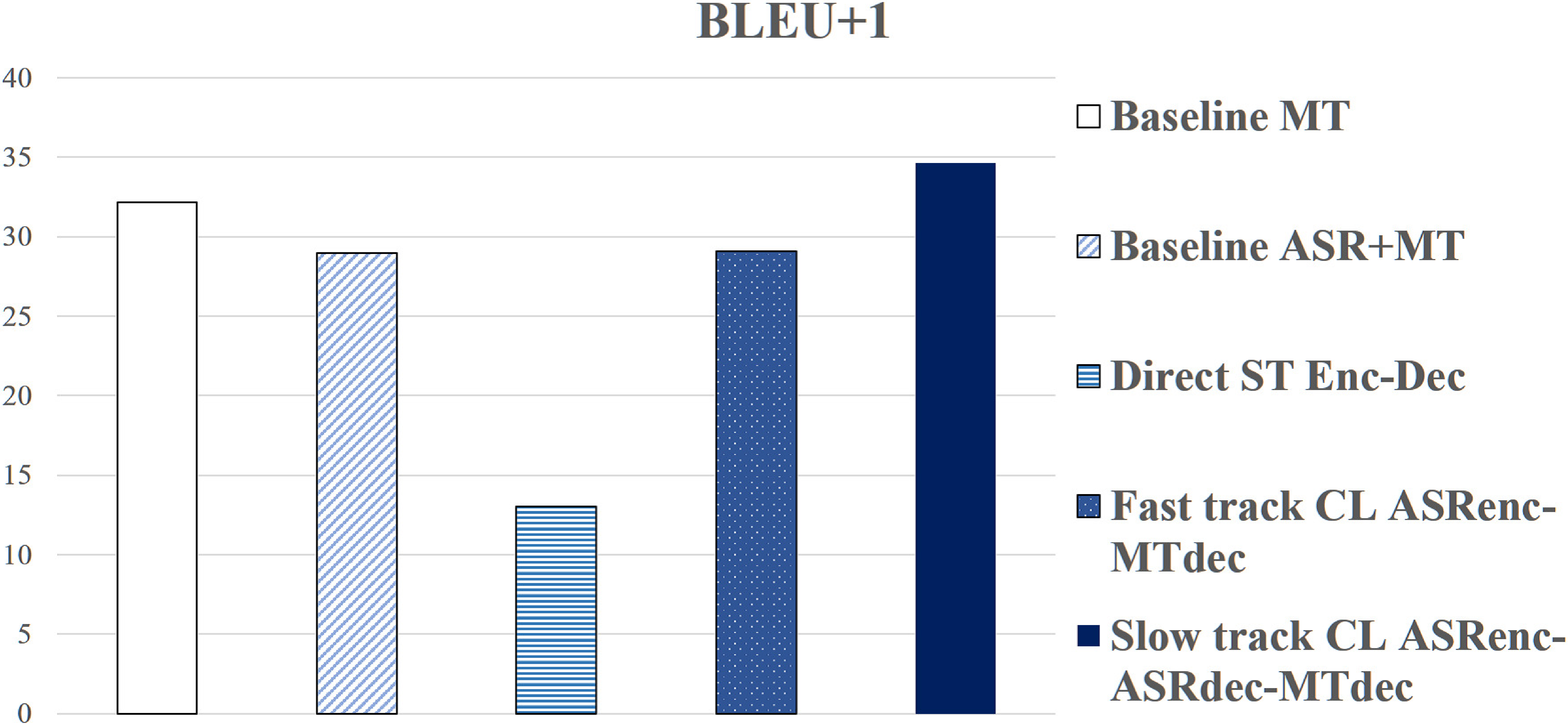}}
  \vspace{-0.3cm}
  \caption{Translation accuracy of each model}
  \label{bleup1}
\end{figure}

First, we show how our proposed methods work during the training steps. Fig.\ref{loss} illustrates the softmax cross-entropy until 15 epochs. The MT system has easiest task, which is translating the text in the source language to the corresponding target language. The loss decreased quite fast. On the other hand, direct ST training is hard, and therefore it gave the worst performance (its loss only decreased $0.04$ from epochs 1 to 15). By using our CL-based proposed method, we can further decrease the loss. Specifically, the one that trained with CL type 1 - Slow Track successfully outperformed the text-based MT system.

Next, we investigated the translation quality of the models summarized in Fig.\ref{bleup1}. The results also reveal that the direct attentional ST system is difficult.
Direct ST Enc-Dec model seems to be over-fitting the language model and could not handle the input speech utterances. The results also demonstrated that our proposed ST Enc-Dec (CL type 1 - Fast Track) model significantly improved the baseline. The best performance was achieved by the proposed ST Enc-Dec (CL type 1 - Slow Track) model, which even surpassed the text-based MT and cascade ASR+MT systems. This system is constructed with [ASRenc+att]+[ASRdec-MTenc]+[MTatt+dec] (Fig.\ref{fig1}). The combination of the second and third parts actually resembles a conventional MT system. Therefore, from the MT system viewpoint, the additional components in the first part, which introduced more noise to the input of the MT system, might function as a denoising encoder-decoder that prevents over-fitting.

\vspace{-0.1cm}
\section{Conclusions}
\vspace{-0.1cm}
In this paper, we achieved English-Japanese end-to-end speech to text translation without being affected by ASR error. Our proposals utilized structured-based CL strategies for training attentional-based ST systems in which we start with the training of attentional ASR and gradually train the network for end-to-end ST tasks by adapting the decoder part. Experimental results demonstrated that the learning model is stable and its translation quality outperformed the standard MT system. The best performance was achieved by our proposed model. Our current results, however, still rely on synthetic data. In the future, we will investigate the effectiveness of our proposed method using natural speech data, investigate various possible language pairs, paralinguistic information, and expand the speech-to-text translation task to a speech-to-speech translation task.

\vspace{-0.1cm}
\section{Acknowledgements}
\vspace{-0.1cm}
Part of this work was supported by JSPS KAKENHI Grant Numbers JP17H06101,JP17H00747 and JP17K00237.

\bibliographystyle{IEEEtran}
\bibliography{mybib}
\end{document}